# Deep Learning in Earthquake Engineering: A Comprehensive Review

Yazhou Xie[*]

Department of Civil Engineering, McGill University, Montreal, QC H3A0C3, Canada

**Abstract**

This article surveys the growing interest in utilizing Deep Learning (DL) as a powerful tool to address challenging problems in earthquake engineering. Despite decades of advancement in domain knowledge, issues such as uncertainty in earthquake occurrence, unpredictable seismic loads, nonlinear structural responses, and community engagement remain difficult to tackle using domain-specific methods. DL offers promising solutions by leveraging its data-driven capacity for nonlinear mapping, sequential data modeling, automatic feature extraction, dimensionality reduction, optimal decision-making, etc. However, the literature lacks a comprehensive review that systematically covers a consistent scope intersecting DL and earthquake engineering. To bridge the gap, the article first discusses methodological advances to elucidate various applicable DL techniques, such as multi-layer perceptron (MLP), convolutional neural network (CNN), recurrent neural network (RNN), generative adversarial network (GAN), autoencoder (AE), transfer learning (TL), reinforcement learning (RL), and graph neural network (GNN). A thorough research landscape is then disclosed by exploring various DL applications across different research topics, including vision-based seismic damage assessment and structural characterization, seismic demand and damage state prediction, seismic response history prediction, regional seismic risk assessment and community resilience, ground motion (GM) for engineering use, seismic response control, and the inverse problem of system/damage identification. Suitable DL techniques for each research topic are identified, emphasizing the preeminence of CNN for vision-based tasks, RNN for sequential data, RL for community resilience, and unsupervised learning for GM analysis. The article also discusses opportunities and challenges for leveraging DL in earthquake engineering research and practice, highlighting the need for open-access multimodal big data and efforts to enhance model interpretability and incorporate physics information into DL. Finally, the paper advocates for DL applications to further advance the research frontier of uncertainty quantification in performance-based earthquake engineering.

## 1. Introduction

While the history of Deep learning (DL) could be traced back several decades in the 1940s with the early development of artificial neural networks (ANN), its renaissance only emerged in the 2010s because of three factors: the availability of large-scale datasets, the significantly increased computing power, and the evolution of algorithmic techniques in network architecture and training scheme. DL's exceptional performance stems from its ability to derive meaningful, high-level features from raw sensory data. This capability is honed through statistical training and learning against large datasets, enabling the creation of an efficient representation of the input space. To date, DL has seen remarkable advancements across various domains, revolutionizing fields such as natural language processing [1], computer vision [2–4], speech recognition [5], medical science [6], game play [7], robotics [8], autonomous vehicles [9], among others.

[*] Email: tim.xie@mcgill.ca

In recent years, earthquake-related fields have also seen a surge of interest in DL applications, which motivated the research community to conduct several relevant state-of-the-art reviews. For instance, Xie et al. (2020) [10] reviewed the promise of implementing machine learning (ML) techniques (e.g., artificial neural network, support vector machine, decision tree, random forest) in four topic areas of earthquake engineering, including seismic hazard analysis, system identification and damage detection, seismic fragility assessment, and structural control for earthquake mitigation. Zhang et al. (2021) [11] presented four DL algorithms, including feedforward neural network (FNN), recurrent neural network (RNN), convolutional neural network (CNN), and generative adversarial network (GAN), and their applications in geotechnical engineering. Avci et al. (2021) [12] focused on vibration-based damage detection in civil structures where autoencoder (AE) and CNN have been pinpointed with predominant implementations. As a data-rich subfield in geoscience, seismology has also benefited from the advancement of DL techniques, which pushed the forefront of research on various seismological tasks [13]. Other related review articles surveyed (1) DL-based computer vision technology in structural damage detection [14], (2) six commonly used DL models in earthquake disaster assessment [15], and (3) the utilization of GAN for the generation of synthetic seismic signals and its applications in seismic-related geophysical studies [16]. Most recently, Cha et al. (2024) [17] also provided a comprehensive review of DL applications in structural health monitoring.

None of the aforementioned review articles covers a consistent research landscape that intersects DL and earthquake engineering. The mismatch lies in either the methods surveyed (e.g., Xie et al. [10] focused on ML techniques; Marano et al. [16] only surveyed GAN) or the field of interest, namely the fields of, e.g., geotechnical engineering [11], seismology [13], and structural health monitoring [17] have overlapping yet distinct DL applications compared to earthquake engineering. In particular, earthquake engineering is an interdisciplinary engineering branch that describes earthquake hazards at the source, characterizes site effects and structural responses, evaluates and mitigates seismic vulnerability and risk, as well as develops seismic codes and standards. The multiple disciplines involved in earthquake engineering motivate the current study to search and select nearly 200 DL-based journal publications from prominent academic databases, such as Web of Science, Engineering Village, and Wiley Online Library. As shown in Figure 1, the overall body of knowledge from these selected journal articles indicates a rapidly increasing research frontier of exploring DL as an innovative tool to deal with various challenging research problems in earthquake engineering. The associated rich body of knowledge warrants a separate, in-depth survey that is currently missing in the literature.

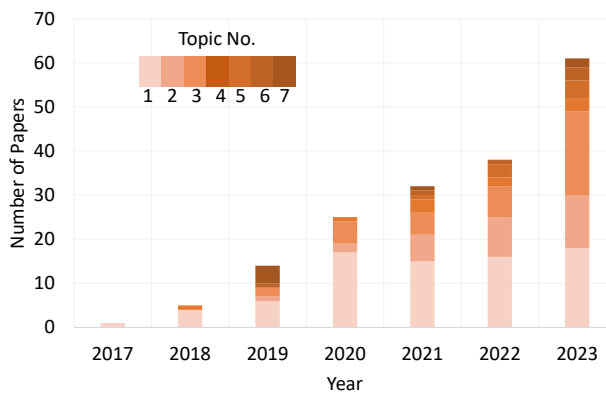

**Figure 1**. The growing interest in DL applications in earthquake engineering

DL applications in earthquake engineering can be grouped into different topic areas depending on the types of data as the inputs and research tasks as the outputs. As illustrated in Figure 2, the main body of the literature can be categorized into seven distinguishable research topics, including (1) vision-based seismic damage assessment and structural characterization, (2) seismic demand and damage state prediction, (3) seismic response history prediction, (4) regional seismic risk assessment and community resilience, (5) ground motion for engineering use, (6) seismic response control, and (7) the inverse problem of system/damage identification under seismic loads. Each research topic has triggered a different number of studies (Figure 1) that applied a variety of DL techniques (Figure 2). It should be noted that these seven research topics do not encompass a handful of extra relevant studies that deal with image data augmentation [18], seismic damage classification through textual description [19], design of structural components [20] and base isolators [21], etc.

Given the overall portrayal of the knowledge structure shown in Figure 2, the current study conducts a comprehensive literature survey of DL applications in earthquake engineering by (1) introducing and discussing the methodological advances for each identified DL technique and (2) disclosing a detailed picture of DL applications in each research topic. The survey is then completed with an in-depth discussion about the specific aspects of earthquake engineering that avail interesting opportunities, as well as pose additional challenges, for DL.

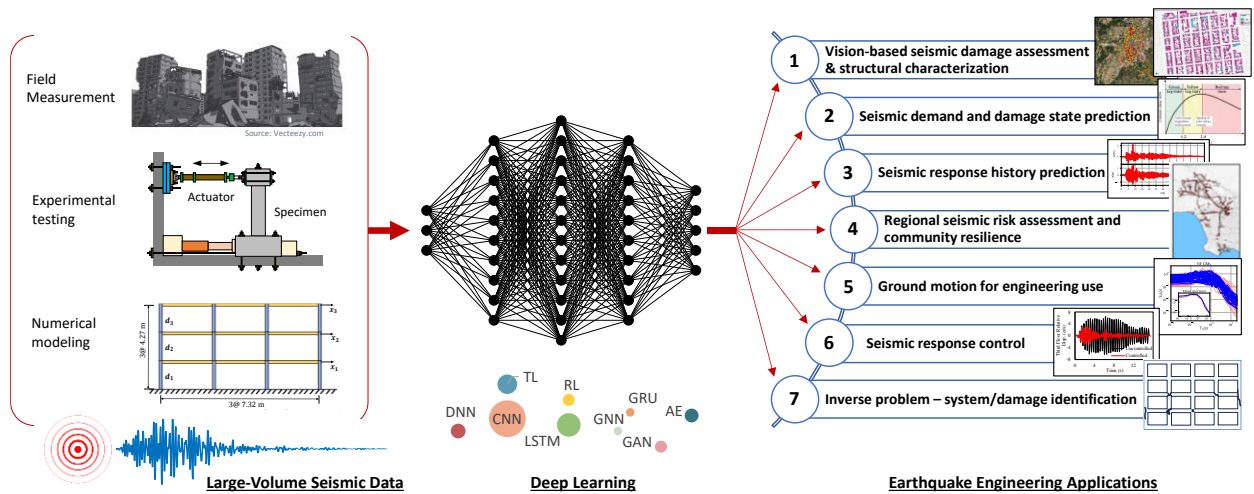

**Figure 2**. DL to deal with different research topics in earthquake engineering (two images adapted from [22,23])

## 2. Deep Learning Techniques Applied in Earthquake Engineering

A variety of DL techniques have been applied to deal with challenging problems in earthquake engineering. Based on whether the training data is labeled, DL techniques can be broadly divided into supervised learning, unsupervised learning, hybrid learning that combines both, and other relevant techniques. As shown in Figure 3, the literature review identifies several DL techniques under each category that have seen a wide spectrum of applications. This section briefly discusses the methodological advances associated with each DL technique, as well as its applications to deal with different research topics mentioned above.

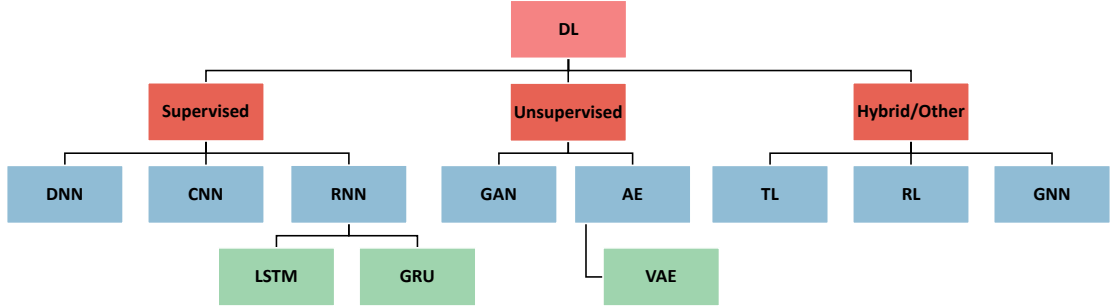

**Figure 3.** Widely applied deep learning techniques in earthquake engineering

## 2.1. Deep neural network - multi-layer perceptron

Some studies use the general terminology deep neural networks (DNN) to refer to all DL algorithms that utilize more than three layers of neural networks. For the purpose of differentiating it from other techniques mentioned below, DNN in this study instead refers to its foundation architecture of multi-layer perceptron (MLP), a supervised learning technique with feedforward networks composed solely of fully connected (FC) layers. As shown in Figure 4, the MLP in DL consists of a number of units (neurons) that are connected by weighted links and organized through an input layer, multiple hidden layers, and an output layer. An example of the computation at the *k*th layer can be written as

$$a_j^{[k]} = f(\sum_{i=1}^{n} W_{ij} \times a_i^{[k-1]} + b) \tag{1}$$

where $W_{ij}$, $a_i^{[k-1]}$, and $a_j^{[k]}$ are the weights, input activations, and output activations, respectively, $b$ is the bias term, and $f(\cdot)$ is the nonlinear activation function that triggers a neuron to generate an output only if the inputs cross a threshold. Various forms of commonly used nonlinear activation functions are also presented in Figure 4. The feedforward networks of MLP ensure that all the computation is performed as a sequence of operations on the outputs of a previous layer, whereas as shown in Equation (1), the FC layers consider that all output activations are composed of a weighted sum of all input activations. The MLP aims to tune the weights in the network such that the network performs a desired mapping of input-to-output activations. The difference between the ideal and current mapping is referred to as the *loss* (*L*). When training the MLP, the weights are usually updated using an optimization process termed gradient descent. An efficient way to compute the partial derivatives of the gradient is through *backpropagation*, which passes values backward through the network to compute how the *loss* is affected by each weight.

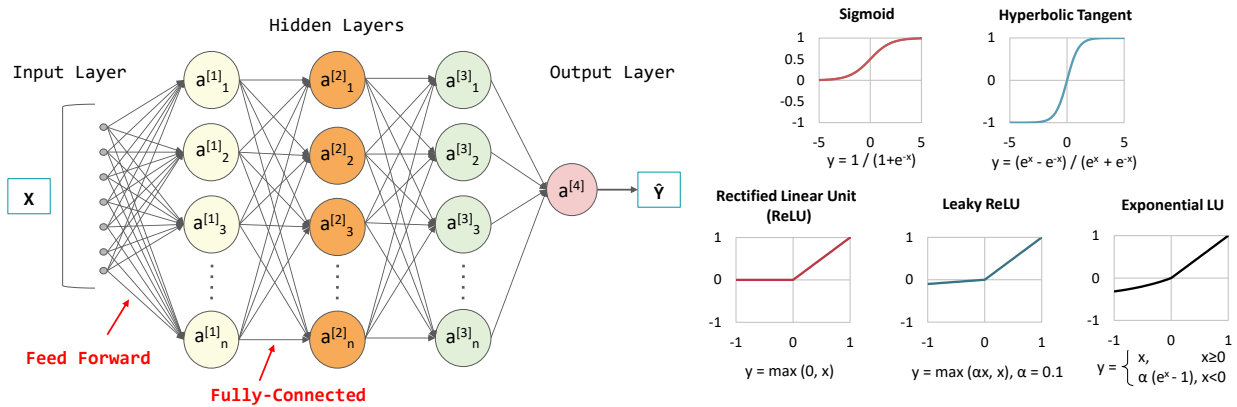

**Figure 4.** Multi-layer perceptron and various forms of nonlinear activation functions

MLP has been applied to address several regression and classification tasks in topic areas such as seismic demand and damage state prediction, regional seismic risk assessment and community resilience, and ground motion for engineering use. While MLP has been found suitable to deal with tabular datasets consisting of scaler inputs, its applicability is somewhat challenged by other more advanced techniques (e.g., CNN and RNN) when dealing with complex tasks that involve other forms of big data, such as image or 2D data, and sequential data.

## 2.2. Convolutional neural network

As an extension of MLP, few modifications have been made in the CNN model architecture, which resulted in a significant breakthrough of CNN in big data applications, particularly in dealing with image data inputs with high accuracy and efficiency (e.g., the ImageNet Challenge [2]). The modifications include the integration of 2D convolution layers, the addition of the pooling layer, and the associated sparse connectivity versus all FC layers to reduce the required amount of storage and computation. Figure 5 shows a typical 2D CNN that is composed of multiple convolutional and pooling layers, each generating a successively higher level abstraction of the input data, termed a *feature map* that preserves essential yet unique information. As shown in Figure 5, a convolution performs dot product operations between the input and a *kernel* (i.e., namely *filter* or *receptive field*) and sums up the resulting products, where the hyperparameter *stride* determines how the *kernel* slides over the input array in both width and height directions. Furthermore, a *pooling* operation is applied to further reduce the dimensions of the input feature map, enabling the network to be robust and invariant to small shifts and distortions. In particular, *pooling* combines a set of values in its receptive field into a smaller number of values through operations such as max, average, and min pooling (Figure 5).

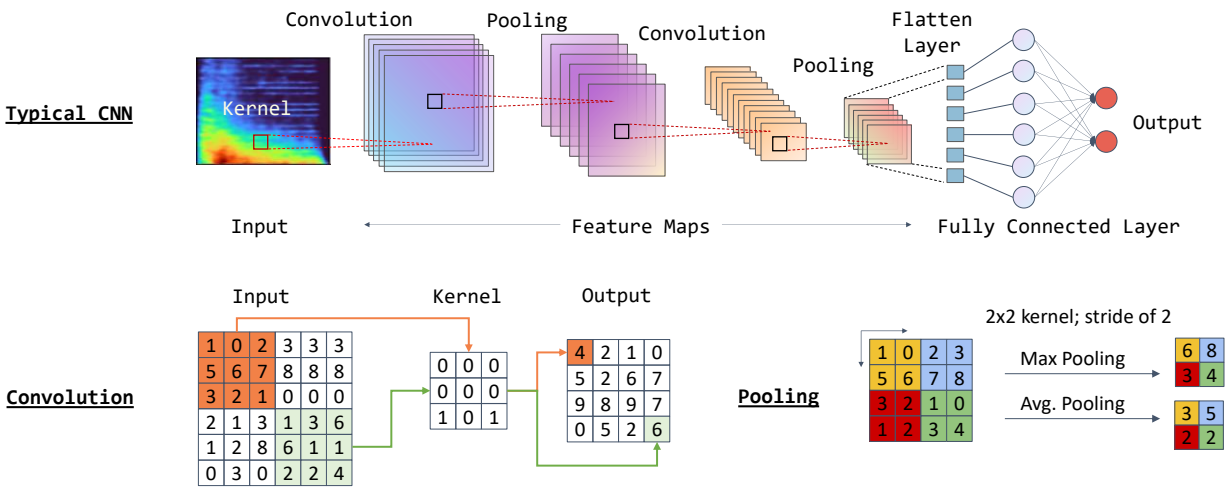

**Figure 5.** A typical 2D Convolutional Neural Network, its convolution operation, and two forms of pooling

CNN bears the capability to automatically extract essential features from the input without the need for human intervention. To date, the computer science community has developed various variants of CNN, including VGGNet [24], MVSNet [25], AlexNet [26], RefineNet [27], RCNN [3], Faster R-CNN [28], Mask R-CNN [29], YOLO [30], U-Net [31], ResNet [32], among others. CNN and its variants have seen predominant applications in vision-based seismic damage assessment and structural characterization. Besides, both 1D and 2D CNN have been implemented in dealing with scaler, sequential, and image data in research topics such as seismic demand and damage state prediction, seismic response history prediction, and the inverse problem of system/damage identification.

## 2.3. Recurrent neural network

RNN is a powerful model for processing sequential data. Figure 6(a) illustrates the computational graph of a representative RNN model; its network architecture is designed to produce an output at each time step with recurrent connections between hidden units. The recurrent formulation is accompanied by the sharing of parameters (e.g., shared weight matrices $U$, $V$, and $W$ in Figure 6(a)), which ensures that each element of the output is produced using the same update rule applied to previous outputs. The parameter sharing also enables generalization in model learning through a few training samples. The RNN is typically trained by the *backpropagation through time* algorithm, which obtains the gradients first on the internal nodes and then on the parameter nodes. In addition to the RNN model shown in Figure 6(a), other important design patterns for recurrent connections include (1) using the output at the previous step and (2) reading an entire sequence to produce a single output [33].

Standard RNN has the issue of vanishing gradients, which makes learning long data sequences challenging. One successful technique for addressing vanishing gradients came in the form of the long short-term memory (LSTM) model developed by Hochreiter and Schmidhuber (1997) [34]. As shown in Figure 6(b), LSTM resembles standard RNN but with each ordinary recurrent node replaced by a *memory cell*, which is equipped with an internal state and three types of multiplicative gates. The forget gate deletes information that is no longer needed; the input gate selects information to add to the cell state; and the output gate decides what information is required for the current hidden state. The values of these three gates are computed through fully connected layers with sigmoid activation functions, ensuring a scaled value range of (0, 1). The input and output gates prevent irrelevant information from entering or leaving the cells, whereas the forget gates enable unbiased and continuous predictions, as they can make cells completely forget their previous states. As such, the vanishing gradient problem encountered by standard RNN is solved through the memory cell by introducing a constant error flow, which enables LSTM to more effectively tackle long data sequence problems.

A streamlined version of LSTM is the gated recurrent unit (GRU) [35], which is also designed to model sequential data by allowing information to be selectively remembered or forgotten over time. GRU is designed with fewer parameters, making it easier and more efficient to train the RNN model. As shown in Figure 6(c), GRU directly uses the hidden state to transfer information through two gates, namely a reset gate and an update gate. The reset gate decides how much past information to forget (i.e., for short-term dependency), whereas the update gate allows long-term control of the data flow in time steps. The simpler gating mechanism of GRU makes it a better candidate for increasing the speed performance of RNN models.

RNN and its variants of LSTM and GRU have been applied to deal with research problems that involve sequential data, where the corresponding research topics include seismic response history prediction, ground motion for engineering use, and seismic response control.

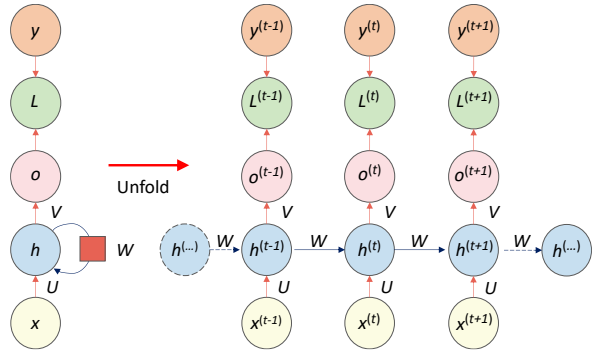
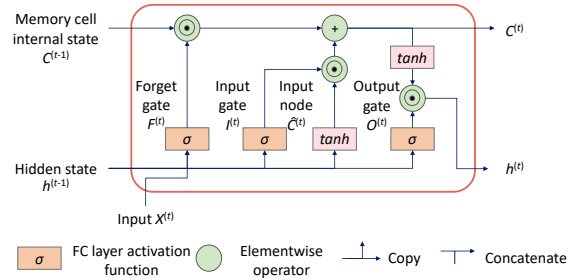

(b) LSTM cell and its operations

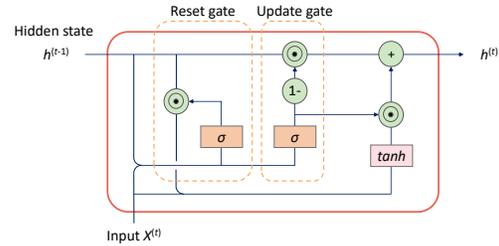

(a) The computational graph computes the training loss of a recurrent network that maps an input sequence of $x$ values to a corresponding sequence of output $o$ values. A loss $L$ measures how far each $o$ is from the corresponding training target $y$. The RNN has input to hidden connections parametrized by a weight matrix $U$, hidden-to-hidden recurrent connections parametrized by a weight matrix $W$, and hidden-to-output connections parametrized by a weight matrix $V$ [33]

(c) GRU cell and its gates

**Figure 6.** A representative example of RNN and its two variants of LSTM and GRU

## 2.4. Generative adversarial network

As initially proposed for unsupervised learning, the generative adversarial network (GAN) was first developed by Goodfellow et al. (2014) [36] as a generative model that consists of two interlinked neural networks. As shown in Figure 7, **G** is the generator that takes random noise in the latent space as input to produce fake samples that are close to real samples. On the contrary, **D** is the discriminator whose role is to differentiate between real and fake samples. GAN is trained in an *adversarial* manner as a supervised learning problem where the generator and discriminator are involved in a minimax two-player game. In particular, GAN is trained for the generator to produce convincing fake data until the discriminator is deceived approximately 50% of the time.

GAN has seen a handful of applications in earthquake engineering to deal with different research topics. As discussed below, these research topics include (1) vision-based seismic damage assessment when subjected to low data and imbalanced class [37], (2) seismic response history prediction for buildings under undamaged and damaged conditions [38], (3) regional seismic risk assessment for generating seismic hazard and damage maps [39], and (4) the generation of artificial ground motions for engineering use [40].

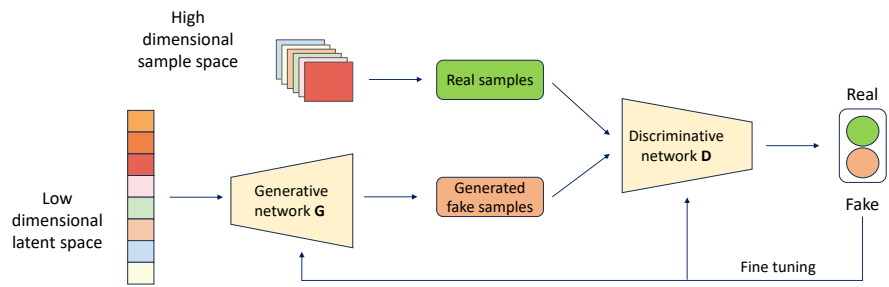

**Figure 7**. Architecture of GAN

## 2.5. Autoencoder

Autoencoder (AE) is another type of generative model used in unsupervised learning [33]. It is based on the idea of finding representation of latent information through two main components: the *encoder* and the *decoder*. As shown in Figure 8(a), the encoder converts the data input into a low-dimensional embedding for a latent representation, whereas the decoder is designed to decompress the low-dimensional embedding and reconstruct the input data. The encoder-decoder architecture essentially learns to compress the data while minimizing the reconstruction error.

Variational autoencoder (VAE) [41] extends the traditional AE architecture by introducing a probabilistic framework for generating the compressed representation of the input data. In particular, traditional AE embeds the input into a vector representation, whereas VAE embeds the input into a pre-defined distribution in the latent space (Figure 8(b)). The latent space in VAE is regularized through KL divergence to have latent variables follow independent standard Gaussian distributions. The regularizer adopted in VAE avoids overfitting and makes latent variables continuous, interpretable, and sometimes linked to certain features of the original data.

The capability of low-dimensional embedding of the original data makes AE and VAE a promising tool to be used in seismic damage detection (e.g., [42]) and ground motion classification and generation (e.g., [43]).

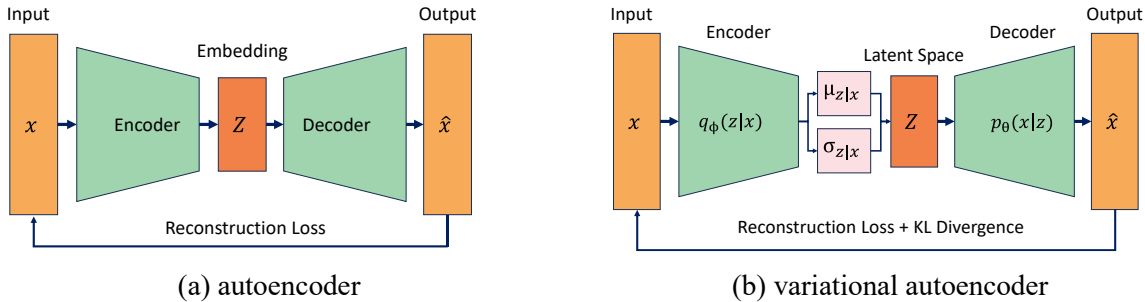

(a) autoencoder    (b) variational autoencoder

**Figure 8**. Architecture of (a) autoencoder and (b) variational autoencoder

## 2.6. Transfer learning

Transfer learning (TL) is a technique that utilizes a DL model previously trained for a given task (source domain) as the base model for refinement against a new task (target domain), where less data is required for the new training. The source and target domains typically share a similar learning task. TL helps quickly tune the parameters for the new task and improves the model performance in generalization, accuracy, and time efficiency. Different TL settings can be defined based on the type of task and the nature of the data available in the source and target domains. In general, TL approaches can be categorized into instance-transfer, feature-representation-transfer, parameter-transfer, and relational-knowledge-transfer [44]. TL has been combined with CNN to deal with several research tasks in vision-based seismic damage assessment (e.g., [45,46]).

## 2.7. Reinforcement learning

Reinforcement learning (RL) is a sub-area of machine learning that deals with sequential decision-making [47]. It devises a framework where a computer agent learns to perform a task through repeated trail and error interactions within a dynamic environment. The standard theory of RL is defined by a Markov Decision Process where (1) the agent chooses an action according to the policy at the current state and (2) the environment receives that action, produces a reward, and transfers to the next state. The objective of RL is to find the optimal policy for the agent to maximize the expected cumulative reward over the future (also termed the value of the state). RL involves managing a significant number of state-action pairs for a track of value (reward) attached to each action, making it challenging to deal with real-world scenarios that involve all different possibilities. In this regard, deep RL can be applied to combine RL with a deep artificial neural network, which is trained to predict the values or evaluate the policies against all possible actions under a given state. Deep RL can be model-based or model-free: the former learns a model to predict the reaction of the environment under a given action, yet the latter directly learns the value and/or the policy from experience (data). Widely used model-free deep RL algorithms include (1) value-based methods (e.g., Deep Q-learning [48]) that update the value function to learn a suitable policy, (2) policy-based methods that learn the policy directly [49], and (3) the actor-critic algorithm that computes the policy gradient using a value-based critic function [50].

In earthquake engineering, deep RL has been primarily implemented in community resilience studies to generate optimal reconstruction plans (e.g., [51]) or select optimal repair decisions (e.g., [52]), which helped communities prioritize improvements to enhance and maximize seismic resilience.

## *2.8. Graph neural network*

Graph neural networks (GNN) are specifically designed neural architectures operated on graph-structure data, which consists of a collection of entities (nodes) with their relationships (edges) represented by a graph. GNN aims to iteratively update the node representations by aggregating the representations of node neighbors and their own representations in the previous iteration. Figure 9(a) illustrates a general design framework for a GNN model [53]. A graph structure needs to be first defined as the model input. The original data can have an explicit (e.g., transportation network) or implicit (e.g., image) graph structure with directed/undirected edges, homogeneous/heterogeneous nodes and edges, and static/dynamic features. A GNN model can then be constructed against a loss function that depends on the training setting (e.g., supervised, unsupervised, or semi-supervised) and the learning tasks at node, edge, or graph level, as illustrated in Figure 9(b). A GNN layer might be designed to have different computational modules, such as propagation, skipping, sampling, and pooling [53]. GNN variants, such as graph convolutional networks, graph attention networks, and graph recurrent networks, have demonstrated ground-breaking performance on many DL tasks that involve graph structures with complex inter-node relationships and interdependency [54].

In earthquake engineering, the field of regional seismic risk assessment and community resilience has seen a couple of GNN applications. One application is for modeling the topology and interdependency of complex infrastructure networks [183], and the other is combined with RL to select optimal repair decisions for water distribution networks [52].

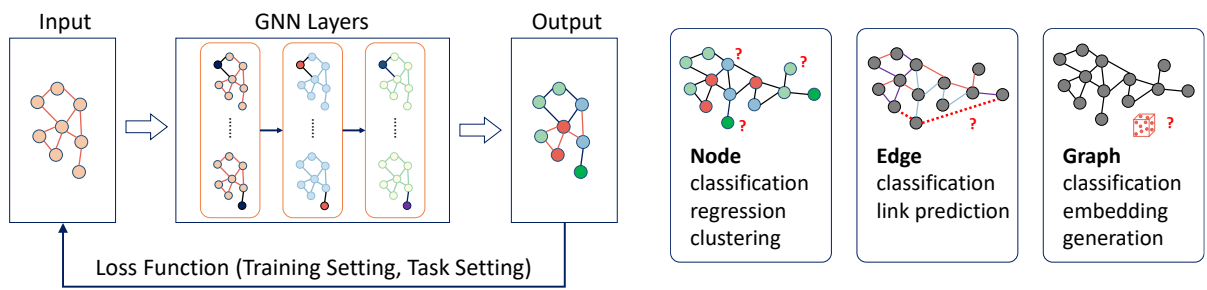

(a) general design framework          (b) graph learning tasks

**Figure 9**. (a) The general design framework and (b) graph learning tasks for GNN

## 3. Deep Learning Applications in Earthquake Engineering

### *3.1. Vision-based seismic damage assessment and structural characterization*

#### *Vision-based seismic damage assessment*

Many studies are motivated by the need for rapid and automated seismic damage assessment across a large region for equipping first responders, decision makers, and stake-holders with an important source of information to enable timely post-earthquake response, loss evaluation, and retrofit planning. The state of research on vision-based seismic damage assessment can be generally classified into two categories. The first is to deal with remote sensing imagery (i.e., satellite, aerial, and, particularly, UAV imagery) to achieve regionwide detection of damaged or collapsed buildings in the wake of an earthquake event (e.g., [45,55], Figure 10(a)). Studies in the second category have focused on handling ground photographs obtained through (1) field surveys during earthquake reconnaissance practice and (2) lab testing of different structural components to accomplish high-resolution seismic damage classification in physical states (e.g., concrete cracking and spalling, rebar exposure, buckling, and fracture) and damage levels (e.g., slight, moderate, extensive, complete) [56–58] (Figure 10(b) and (c)). Other than these two tasks, a handful of studies have also considered leveraging image data to distinguish damage caused by corrosion versus earthquake [46], predict stiffness and strength degradation [59], identify beam deflection [60], and detect earthquake-induced secondary hazards, such as landslide [61] and ground failure [62].

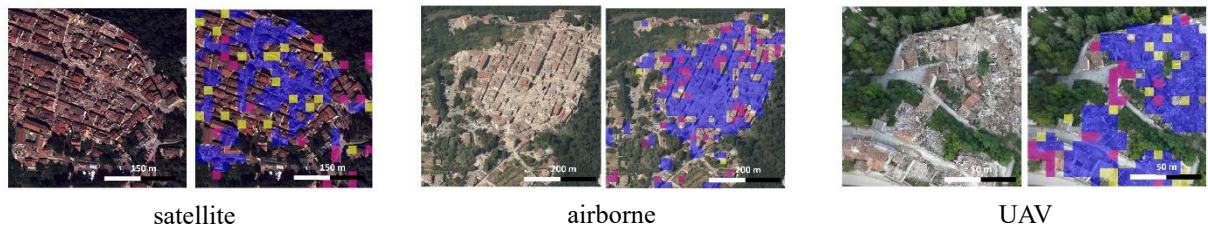

(a) building damage detection using remote sensing imagery (blue – detected damage, violet – false positive, yellow – false negative) (adapted from [63])

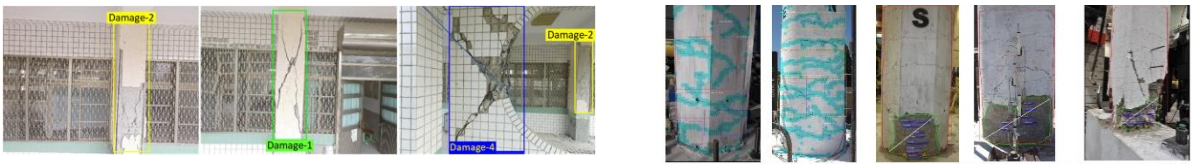

(b) damage level classification for earthquake reconnaissance (Damage 1 – surface crack, Damage 2 – spalling, Damage 3 – spalling with exposed rebars, and Damage 4 – severely buckled rebars) (adapted from [58])

(c) damage assessment for experimentally tested columns (Damage Sate 2 – left two, Damage State 4 – $3^{rd}$ and $4^{th}$, Damage Sate 5 – last one) (adapted from [56])

**Figure 10.** Example studies on vision-based seismic damage assessment

CNN has been considered in many studies as the most effective DL tool to automatically extract strong deep features for image classification and semantic segmentation, where widely-applied CNN-based models include ResNet [64], U-Net [65], YOLO [66], Faster R-CNN [62], Mask R-CNN [67], RefineNet [68], among others (Figure 11). CNN has also been combined with TL (e.g., [45,46,63,69]) to improve model transferability and generalization against different datasets. The model performance of these DL frameworks is commonly evaluated through confusion matrix (accuracy, recall, precision) and its derived metrics such as overall accuracy (OA), kappa coefficient, F1 score for damage detection and classification, as well as the intersection over union (IoU) that quantifies the degree of overlap between two image boxes for damage segmentation (Figure 11).

A detailed review identifies several representative studies in the literature. The methodological advancement from these studies include (1) the combination of aerial images and 3D point cloud features to improve CNN model accuracy [103]; (2) the use of a RNN as a probabilistic anomaly detector on the temporal changes of synthetic aperture radar (SAR) to estimate earthquake damage [100]; (3) the application of TL to achieve structural component characterization, damage level classification, and damage type determination [77]; (4) a convolutional AE model with flexible configurations for pixel-level recognition of damaged buildings under diverse weather conditions [106]; (5) a novel loss function fusing geometric consistency constraint with cross-entropy loss in a U-Net framework for post-earthquake building segmentation with complex geometric features across multiple scales [65]; (6) a three-level CNN approach with Bayesian optimization for hyperparameter selection for bridge system-level failure classification, column detection, and damage localization [85]; (7) an integration of the DeepLab framework with the linear iterative cluster method and mathematical morphological approach to obtain clearer boundaries for damaged building areas [99]; (8) a multimodal and multitemporal optical/SAR dataset for building damage mapping under earthquakes, tsunamis, and typhoons [70]; (9) a novel quantum CNN approach to detect multiclass seismic damage of RC buildings from images [73]; (10) the balanced semi-supervised GAN to resolve the low-data and imbalanced-class issue [37]; (11) a combination of 3D point cloud and 2D image for 3D damage quantification [83]; and (12) a multi-task learning approach that simultaneously accomplishes the semantic segmentation of seven-type structural components, three-type seismic damage, and four-type deterioration states [57].

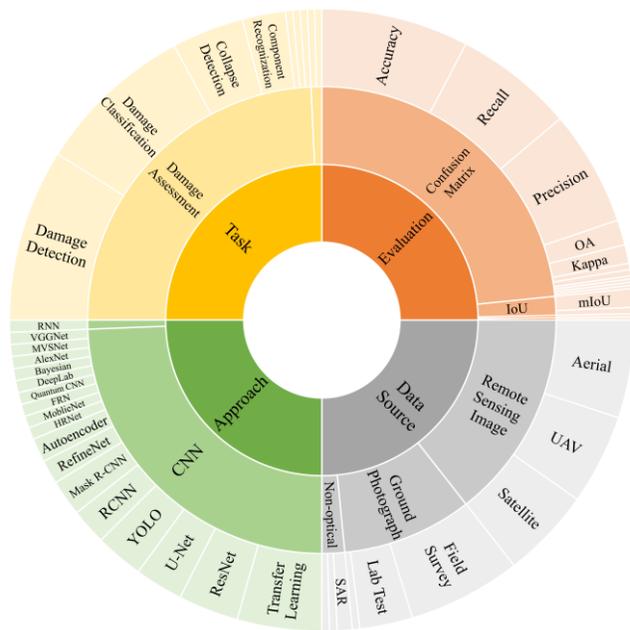

**Figure 11.** Hierarchical distribution of studies for vision-based seismic damage assessment [37,45,46,55–116] according to data source, DL approach, task topic, and evaluation metric

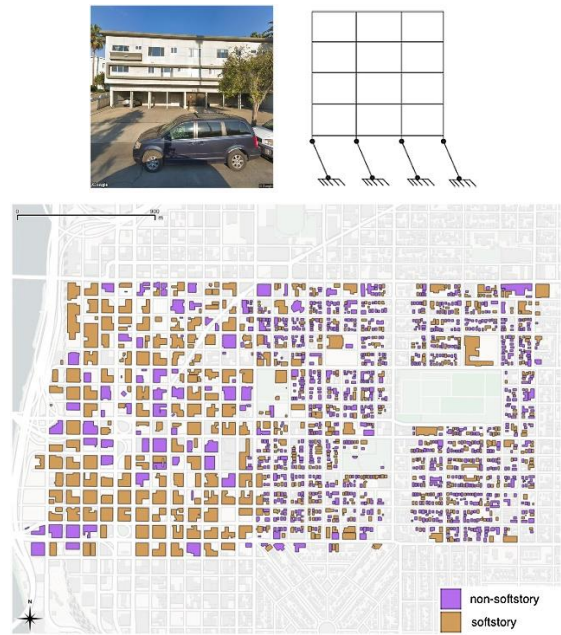

**Figure 12.** Soft-story building classification from street view images for Buckman in Portland, Oregon, United States (adapted from [23])

*Vision-based structural characterization*

Other than rapidly assessing structural damage after earthquake events, a dozen studies have made attempts to handle cite-scale satellite and street view images for characterizing structural systems to support the simulation of earthquake impacts and decision-making, as well as detecting potential structural defects that are vulnerable to earthquake loading [117]. The DL involved is to deal with image-based structural system classification or pixel-level semantic segmentation that extracts certain structural components; all studies have shown the promise of utilizing CNN and its variants, such as Mask R-CNN [118], VGG [119], SegNet [120], to achieve good model performance, which has been commonly evaluated through confusion matrix and F1 score. Street view images are the primary type of image data well suited for DL to identify soft-story buildings [23,121,122] (Figure 12), lateral-load resistant systems in material and type [123,124], masonry buildings [118,119], and bridge components (e.g., columns, beams, and slabs) [120] and substructure types (frame bent, wall, hammerhead wall) [125]. Satellite images have also been utilized to classify the shapes of building roofs [117]. A somewhat unique attempt is to classify soils into different site classes using images designed as a combination of the topographic slope and the mean horizontal-to-vertical spectral ratio (HVSR) of earthquake recordings [126]. Instead of dealing with street view images of building facades, which might be blocked in raw images, Chen et al. [127] segmented 3D point-cloud data in a city, extracted point density features for buildings, and identified soft-story buildings using CNN models.

## *3.2. Seismic demand and damage state prediction*

The studies compiled in this section focus on dealing with the significant computational challenge in conducting nonlinear response history analyses (NRHA) of complex structural systems. DL models are trained as surrogate models to predict the *peak values* from these analyses as seismic demands (e.g., [128,129]), which are further utilized for damage state classification (e.g., [130,131]) and seismic fragility assessment (e.g., [132,133]). As mainly targeting classes of structures for regional assessment, the devised DL algorithms handle data inputs (i.e., those with varying values) from both GMs and structural attributes. Figure 13 presents a chord diagram that summarizes the developed DL models with their applications to address various types of data inputs. For GMs, viable input data formats include (1) the original time-series signal handled by 1D CNN [128,134] and LSTM [135–137]; (2) feature parameters related to earthquake (e.g., magnitude, epicenter distance) and intensity (e.g., peak ground acceleration, response spectra acceleration) as scaler inputs for CNN [138,139], DNN [133,140,141], and Bayesian DL [142]; (3) the spectrogram image processed through Duhamel integral [143], short-time Fourier transform (STFT), and wavelet transform [130,131] to train the 2D CNN while accounting for temporal/spectral non-stationarity of GM; and (4) the spectrum curve in the frequency domain (e.g., pseudo-acceleration spectrum [144] and Fourier spectrum [145]) to train CNN and DNN models (Figure 13). Other than GMs, structural attributes have been converted into data inputs in two ways. The first is to deal with one specific class of structures where its peak responses are considered to be affected by influential feature parameters such as story number [136], natural period [146], material property [147], etc. In contrast, the second approach considers using cyclic hysteresis loops or monotonic force-displacement curves from pushover analyses [129,142,148,149] to more sophisticatedly represent structural behaviors and train DL models.

As is typically phrased to replace NRHA, seismic demands and damage states as data outputs for labeling and supervised learning have been obtained through numerical simulations. The performance of DL models is evaluated through either the confusion matrix for damage state classification or regression metrics, e.g., mean absolute error (MAE), mean square error (MSE), and $R^2$, for seismic demand prediction.

A detailed review also identifies several unique and representative studies in the literature. The methodological advancement from these studies includes (1) the stacked LSTM with specific network architectures for improved accuracy and reduced training time [136]; (2) an AE model for seismic response reconstruction where reconstruction errors are used to detect seismic damage [42]; (3) the use of floor acceleration response history as data inputs to estimate inter-story drift and remaining stiffness ratio [150]; (4) the Bayesian DL with a loss function proportional to the negative log likelihood of the Gaussian distribution function to obtain the mean and variance of structural responses [142]; (5) the utilization of flexibility matrix to represent the target structure where the difference between the two matrices is used to identify seismic damage in near-real-time [151]; (6) the DNN prediction of modal contribution coefficients to improve the accuracy of the response spectrum method [152]; and (7) the training of DNN to predict high-fidelity results from low-fidelity simulations [132].

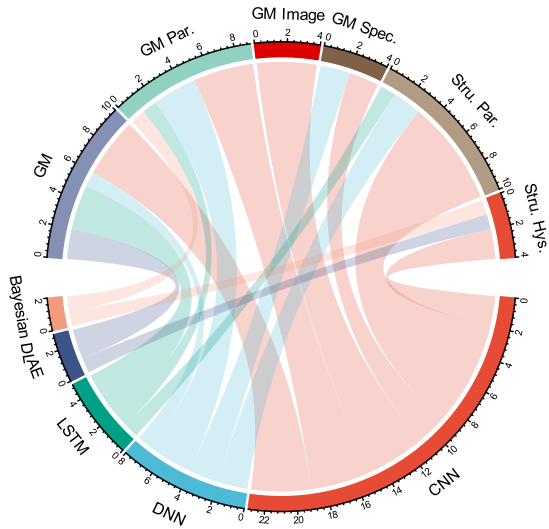

**Figure 13.** DL models to handle different types of data inputs for seismic demand and damage state prediction (Nomenclature: GM – ground motion, Par. – parameter, Spec. – spectrum, Stru. – structural, Hys. – hysteresis) [42,128–159]

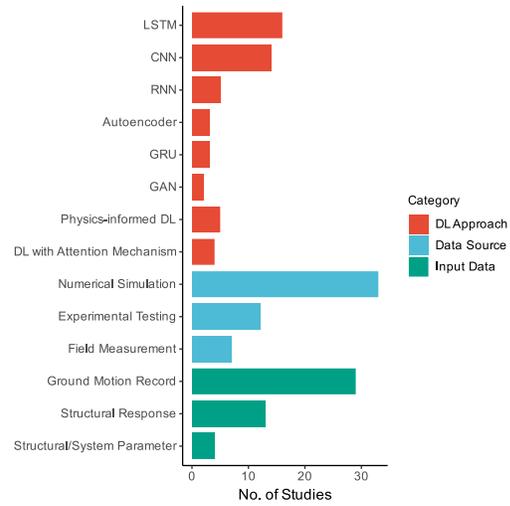

**Figure 14.** DL approach to handle different data inputs from three data sources for seismic response history prediction [38,160–200]

### 3.3. Seismic response history prediction

A more rigorous attempt to leverage DL as a surrogate model is to predict the entire *time history of structural responses* during earthquake events. Figure 14 summarizes the state of research for seismic response predictions in terms of the utilized DL models, their compatible data inputs, and the ways to generate/obtain such data inputs and outputs. Most studies handle model inputs as sequential regression problems to predict time-series response histories of different structural parameters, including inter-story drift, floor acceleration, restoring force, etc., which are primarily generated through numerical simulations (Figure 14). Although CNN is recognized for its capability in the classification of data with grid-like topology (e.g., 2D images), 1D CNN has been applied to deal with GM inputs and structural response outputs with time-sequence structures for extracting the deep features and describing the input-output relationships (e.g., [174,180]). In contrast, as designed to learn sequential, time-varying, linear/nonlinear patterns for regression problems, RNN (e.g., [197]), with specific model architecture such as GRU (e.g., [172]) and LSTM (e.g., [160]), have shown the promise on seismic response modeling. In particular, LSTM is favored in the literature with the highest number of applications (Figure 14) due to its added ability to capture long-range data dependencies for predicting time-series responses.

Many studies are motivated to achieve a complete replacement of finite element modeling that predicts seismic response histories using GMs as model inputs. Methodological advancements identified in the relevant literature include (1) directly integrating DL into the time-stepping numerical integration schemes to solve nonlinear dynamics [170,179,187]; (2) embedding physics constraints in the loss functions for physics-informed DL with enhanced model performance [173,178,189,190]; (3) introducing the attention mechanism that selects dominant features for a higher prediction accuracy [175,177]; (4) addressing structural portfolios where structural attributes are also included as feature parameters to train DL models [162,163,169,175]; and (5) comparing the model performance of LSTM, WaveNet, and 2D CNN in accuracy, efficiency, and robustness [200].

The other research area deals with seismic structural health monitoring (SHM) that utilizes existing structural responses (e.g., those obtained from shaking table tests and field measurements) as data inputs to predict other response quantities of interest, forecast future responses, and pinpoint real-time damage conditions. For instance, Park et al. [196] trained CNNs to estimate strain responses of structural members from measured accelerations and displacements to support long-term monitoring in case of the absence or defect of strain sensors. Pan et al. [165] addressed the limited sensor problem to predict bridge displacement responses at unmeasured locations from those at measured locations. In a similar context, unsupervised DL models with encoder-decoder networks have been developed to (1) provide seismic damage index with a measure of uncertainty at each time instant [168], (2) forecast upcoming responses through historical measurements [199], and (3) reconstruct structural responses in a healthy state and identify the seismic damage based on the difference between input and reconstructed data [171]. Likewise, transient dynamic response and damage classification were achieved by developing a GAN to predict building responses under undamaged and damaged conditions [38].

### 3.4. Regional seismic risk assessment and community resilience

Studies in this section deal with regional seismic risk assessment (RSRiA) and community resilience. RSRiA quantifies regional seismic impacts by convolving four sequential analysis modules: regional seismic hazard analysis, exposure modeling, damage and fragility assessment, and loss assessment [201], whereas community resilience studies further characterize post-earthquake function recovery for regional seismic resilience assessment (RSReA), as well as conduct restoration planning and optimization that help communities prioritize improvements to enhance/maximize seismic resilience [202]. RSRiA and RSReA holistically integrate various underlying physics processes and uncertainties that often require conducting a large number of Monte-Carlo simulations to deal with numerous interconnected structures. In this regard, DL has been applied as surrogate models at different layers to reduce the associated computational burden.

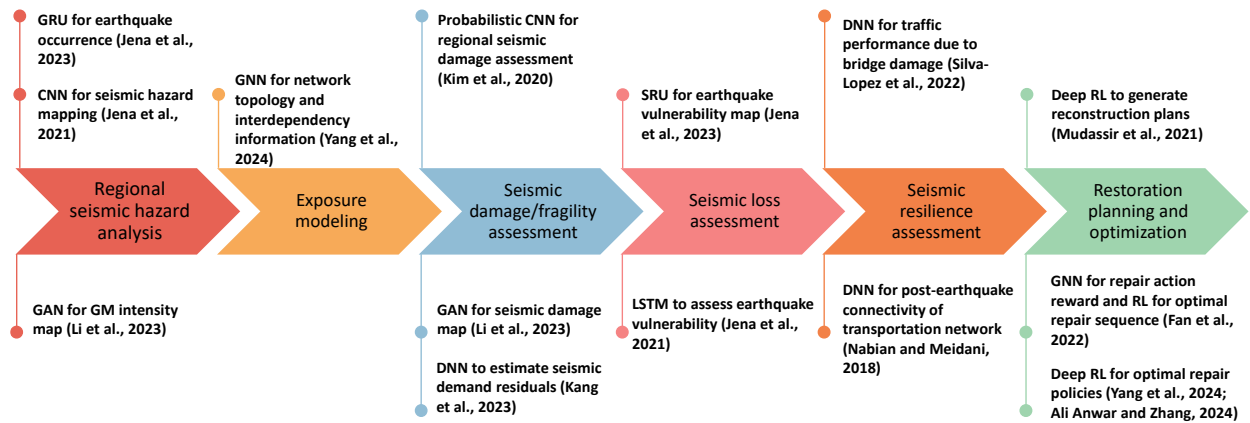

**Figure 15**. DL applications in regional seismic risk assessment and resilient planning [39,51,52,139,203–210]

As summarized in Figure 15, DL applications spread the entire spectrum of analysis modules for RSRiA and community resilience. In regional seismic hazard analysis, GRU has been applied to estimate the spatial probability of earthquake occurrence [210], whereas CNN [208] and GAN [39] have been utilized to develop surrogate models that directly generate GM intensity maps given earthquake source parameters. Exposure modeling poses challenges to capturing the topology and interdependency information of complex infrastructure networks, which has been addressed by developing a GNN model [203]. Seismic damage and fragility assessment involve finite element modeling or nonlinear analysis procedures to obtain probabilistic seismic demand models of different structural portfolios that vary in geometry, material property, design detail, etc. The time-consuming modeling and analysis efforts are dealt with by developing (1) a probabilistic CNN model that estimates area-wide structural damage given the spatial distribution of seismic intensity levels [206]; (2) a GAN-based seismic damage map under earthquake source inputs [39]; and (3) a DNN to estimate the variances and correlations of residuals in engineering demand parameters [89]. For seismic loss assessment, seismic vulnerability maps have been generated using surrogate models developed by SRU and LSTM that consider social, structural, and geotechnical factors [207,210]. RSReA further examines the functionality states of structural systems, where DNNs have been applied to more rapidly predict (1) the changes in traffic performance metrics due to bridge damage [209] and (2) post-earthquake connectivity of a transportation network [205]. To further support decision-making for community resilience, a deep RL algorithm has been utilized to generate reconstruction plans by considering the resources available and the needs of various stakeholders [51]. A convolutional GNN-integrated deep RL model has been developed to select optimal repair decisions to maximize the seismic resilience of water distribution networks [52]. RL has also been applied to train neural networks that (1) approximate the optimal repair policies by considering uncertainties of the restoration process [203] and (2) simultaneously reduce retrofit costs and uncertain future consequences given seismic hazards [204].

### *3.5. Ground motion for engineering use*

GM data has been used in various ways in earthquake engineering, ranging from ground motion models for probabilistic hazard analysis and design spectra definition, to regional hazard simulation and validation, time-series data as inputs for response history analysis and shaking table testing, and intensity measures for constituting fragility curves, etc. Several attempts have been made to leverage DL to deal with GM data that better serves its essential role in earthquake engineering research and practice (Figure 16). Dupuis et al. [211] trained a DNN to estimate the quality (i.e., a score factor determined by high-frequency noise, seismic waveform completeness, the slope and shape of the Fourier amplitude spectrum, instrument malfunction, and records with multiple earthquakes) and minimum usable frequency of GM records from earthquakes in New Zealand. The study is helpful for quality screening of GMs that will be particularly acute for the development of empirical ground-motion models and validation of physics-based ground-motion simulations. Zhao et al. [212] trained a Siamese CNN to select GMs that have the mean response spectrum matches the target spectrum in all periods, whereas Matinfar et al. [213] and Matsumoto et al. [40] trained a GAN to generate artificial spectrum-compatible GMs with matching temporal and frequency characteristics. Moreover, Ning and Xie [43] developed an analysis workflow that transforms and reconstructs GMs through STFT, encodes and decodes their latent features through convolutional VAE, and classifies and generates GMs by grouping and interpolating latent variables. Their study indicated that using five classified, top-ranked motions, regardless of recorded or simulated accelerograms, can achieve reasonable and efficient fragility estimates compared to the case that adopts 230 GMs. Bond et al. [214] applied a similar AE approach for ground motion spectra clustering and selection. A couple of studies have also focused on handling GMs with mainshocks and aftershocks. Ding et al. [215] developed DNN and GAN to predict the spectral accelerations (Sa) of aftershocks using eight seismic variables and Sa of the mainshock at 21 periods as the model inputs. Fayaz and Galasso [216] used LSTM to develop a generalized ground-motion model that estimates consistent vectors of intensity measures for mainshocks and aftershocks. A somewhat unique attempt was to deal with GMs recorded through smart devices – an RNN was developed to detect and remove sliding motions in acceleration measurements from smart devices to reproduce realistic seismic shaking in structures [217].

| Quality assessment | GM selection | GM generation | Mainshock-aftershock | GM from smart device |
|---|---|---|---|---|
| • DNN (Dupuis et al., 2023) | • CNN (Zhao et al., 2022)<br>• Autoencoder (Ning and Xie, 2023; Bond et al., 2024) | • GAN (Matinfar et al., 2023)<br>• GAN (Matsumoto et al., 2023)<br>• Autoencoder (Ning and Xie, 2023) | • DNN, GAN (Ding et al., 2021)<br>• LSTM (Fayaz and Galasso, 2022) | • RNN (Na et al., 2022) |

**Figure 16.** DL to deal with ground motions for engineering use [40,43,211–217]

### *3.6. Seismic response control*

Studies on structural control algorithms to intelligently *control* and mitigate seismic vibrations of structures in real-time have once attracted significant attention in the earthquake engineering community.

In this field, DL has shown its applications to be embedded into structural control frameworks through two modules. One is as a surrogate model, an emulator that predicts structural response history based on the previous response, control signal, and GM input. DL models as structural surrogates include an RNN [218] and an LSTM [219]. The other module is to train a DL-based controller that generates the new control signal (or input force to activate a protective device) in real-time to achieve the desired structural response. Such DL-based controllers have been trained through RNN [218] and RL [220,221], which determines the optimal control force to maximize a reward function that reduces both structural responses and the required control energy. To further enhance the efficiency and robustness of the control system, studies have also focused on developing (1) LSTM-based RNN as a data-driven controller for rapid online adaptation and processing, and pure online learning that minimizes reliance on pre-training [222] and (2) LSTM-based decentralized control method for high-rise buildings to reduce control complexity and improve fault tolerance [223]. In a separate study, LSTM was also used to classify earthquakes to determine the optimal control parameters for a semi-active variable stiffness isolation system [224].

### 3.7. Inverse problem – system/damage identification

Studies compiled in this section deal with the inverse problem where DL is trained against seismic response measurements to identify the structural system or detect/classify seismic damage. CNN has shown more applications than other DL methods (e.g., [225]). Two studies indicated success in identifying system models and predicting component properties. Impraimakis [226] trained CNN to automatically detect the model class based on measured responses without requiring the input information from the model. Zeng et al. [227] developed a DNN and combined it with wavelet packet decomposition to simultaneously monitor axial pressure and shear deformation for laminated rubber bearings. More studies proposed DL approaches to train structural responses under different damage conditions and utilized the pre-trained model to identify or classify seismic damage states. Zhou et al. [228] trained an AE model based on hysteresis loop analysis to predict the seismic damage index associated with stiffness degradation. Yu et al. [229] proposed a CNN method to deal with FFT data from raw sensor measurements and identify local damage for buildings installed with smart base isolators. Khodabandehlou et al. [230] developed a CNN to classify the seismic damage state of a reinforced concrete bridge using a limited number of acceleration measurements from the shake-table test. Lei et al. [231] detected structural damage subject to unknown seismic shaking by coupling the CNN with wavelet-based transmissibility data that eliminates the influence of different seismic excitations.

## 4. Challenges and Opportunities

The numerous DL applications reviewed indicate a significant surge of interest in the community in leveraging this advanced data-driven tool to deal with challenging problems in earthquake engineering. Over the decades, many of these problems have proven difficult to address despite the wealth of domain knowledge, primarily due to factors such as the uncertainty in earthquake occurrence, the unpredictable nature of the seismic load, the nonlinear structural response, the complex interdependency of the built environment, and the imperative of engaging communities for risk reduction and post-earthquake recovery, among others. Moving forward, DL is expected to be utilized as an essential tool for solving these problems, leveraging its data-driven capacity to achieve nonlinear mapping, sequential data modeling, automatic feature extraction, dimensionality reduction, optimal decision-making, etc.

This review paper indicated a viable pathway showing which DL technique is appropriate for which category of problems. For instance, CNN and its variants should be considered with top priority to solve vision-related tasks such as vision-based seismic damage assessment and structural characterization (Figure 11). Conversely, RNN, especially when employing LSTM units, is advantageous for dealing with temporal relations in the data, making it well-suited for research tasks such as seismic response history prediction (Figure 14) and real-time seismic response control (Section 3.6). Tailored for sequential decision-making, RL finds its promising application domain in community resilience studies, encompassing post-earthquake emergency response, recovery planning, and optimization (Figure 15). Unsupervised DL, such as GNN and AE, are fertile with opportunities to deal with GMs for engineering use, where one example would be to generate artificial GMs, as shown in Figure 16.

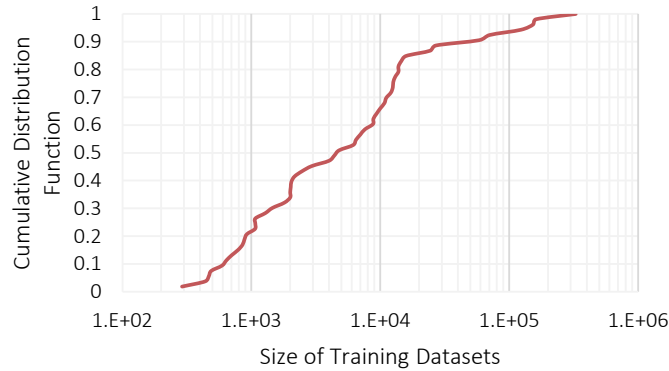

**Figure 17.** Cumulative distribution function for the size of training data used for vision-based seismic damage assessment

However, the pursuit of DL in earthquake engineering does not come without challenges. First, DL is a data-hungry technique, most often requiring the availability of massive, high-quality, labeled (for supervised learning) datasets to ensure its superior performance. Figure 17 indicates the distribution of training data sizes for DL applications in vision-based seismic damage assessment – the required quantity of training data varies significantly, ranging from hundreds to thousands of hundreds, contingent upon the type, scale, and resolution of the image data. Acquiring a sizable dataset may sometimes become intractable when dealing with real-world earthquake events because large and damaging earthquakes are (fortunately) rare. Also, a relevant concern would arise from the data unbalance issue – instances of seismic damage/collapse would be scarce as opposed to undamaged cases. The skewed data distribution not only complicates the DL training process, potentially leading to an overfitted model, but may also undermine the utilization of the DL model as, for instance, the cost for a false negative of true seismic damage might be high. In this regard, efforts should be made to synchronize the advancement of applicable DL techniques with the creation of accessible open-source datasets (e.g., [232]). One solution is to leverage data fusion techniques to develop multimodal, multitemporal, and heterogeneous datasets composed of different resolutions, data types (image, video, text, tabular, and sequential data), earthquake events, seismic regions, data sources (experimental testing, remote sensing, field measurement, and numerical simulation), and structural and infrastructure systems [70]. Accordingly, cutting-edge DL techniques such as TL [44], diffusion models [233], attention mechanisms [234], ensemble learning [235], self-supervised learning [236], and transformer models [237] can be explored to achieve a unified representation of complex big data for different research tasks in earthquake engineering.

The understanding of the underlying physics has been considered arguably the most essential weapon for the community to address various earthquake engineering problems. Conversely, the black-box nature of DL models might cause an atmosphere of skepticism regarding their interpretability, reliability, robustness, and generalizability. DL applications in earthquake engineering are often safety-critical, making it crucial to interpret, understand, and trust the model's decision-making process. While the interpretability of DL raises a concern in some relevant studies, no work has yet made efforts to explain the model's learned parameters and weights, as well as understand the model behavior in depth. In this regard, different interpretability approaches, e.g., gradient-based attribution methods [238,239], guided back-propagation [240], concept activation vectors [241], deconvolution [242], class activation maps [243], layer-wise relevance propagation [244], etc., can be applied along with domain knowledge towards explainable DL for earthquake engineering.

One more rigorous attempt is to integrate physics information into the development of DL models. Table 1 lists relevant studies in earthquake engineering that develop physics-informed DL models. These studies focus on the prediction of seismic response histories of structural systems, where physical laws are obtained from the equations of motion (EOM) governing structures' dynamics equilibriums. Therefore, physical constraints are introduced into the model training process by (1) implicitly incorporating physics loss into the loss function (Table 1), (2) embedding DL into the time-stepping integration method to solve the EOM [170,187], and (3) employing a GNN to represent the physical system (i.e., graph nodes for mass nodes, graph edges for dynamic interactions between nodes, and the adjacency matrix for the physical topology) [195]. While the embedding of physics into DL bears the potential to alleviate overfitting, reduce the need for big data, as well as improve the model robustness, accuracy, and extrapolation ability [190], these studies also underscore the premise that the seismic behavior of a structure needs to be fully characterized through an equivalent yet simplified multi-degrees of freedom (MDOF) system. This assumption poses a challenge when confronted with real-world structural systems, which may comprise numerous components forming a high-dimensional parameter space with complex topology, dynamics, and nonlinearity. Other than seismic response history prediction, the application of physics-informed DL is rarely observed in addressing other subjects in earthquake engineering.

**Table 1**. Applications of physics-informed DL in earthquake engineering

| Paper | Research Topic Number [a] | DL Model | Physics Information |
|---|---|---|---|
| Xiong et al. [173] | 3 | CNN | Loss Function |
| Liu et al. [178] | 3 | LSTM | Loss Function |
| Guo et al. [170] | 3 | Residual Network | Time-step integration |
| Zhang et al. [190] | 3 | LSTM | Loss Function |
| Eshkevari et al. [187] | 3 | RNN | Time-step integration |
| Zhang et al. [189] | 3 | CNN | Loss Function |
| Chen et al. [195] | 3 | GNN | Graph Node and Edge |
| Chou et al. [150] | 2, 3 | LSTM, CNN | Loss Function |

[a] Research topic number is shown in Figure 2.

As the earthquake engineering community is moving away from relying on prescriptive approaches to increasingly embracing a probabilistic way of thinking [201], one important area for DL applications that remains relatively underexplored lies in uncertainty quantification. The literature review identifies a handful of relevant studies. Eltouny and Liang [168] presented an uncertainty-aware early warning system that can provide near real-time SHM under seismic hazards. Stephenson et al. [100] applied RNN to forecast a probability distribution of the coherence between pre- and post-event SAR images, providing a measure of confidence in the identification of seismic damage. Noureldin et al. [141] proposed ensemble probabilistic DL models toward quality-driven prediction intervals of seismic responses for low- to mid-rise buildings with limited irregularity. Kim et al. [142] adopted Bayesian DL for predicting both the mean and variance of seismic responses. Ali Anwar and Zhang [204] integrated Deep RL into a risk optimization framework to simultaneously reduce retrofit costs and uncertain future consequences given seismic hazards. Ning and Xie [43] targeted uncertainties in GMs; they devised a VAE model to classify and select a small set of motions that achieve reliable and efficient seismic fragility assessment. Feng et al. [132] trained a DNN model to predict the probabilistic distribution of high-fidelity results from low-fidelity simulations. These studies should serve as a starting point to advocate for additional endeavors that can further advance the research frontier of uncertainty quantification in earthquake engineering.

## 5. Conclusion

This paper conducts a comprehensive literature survey of Deep Learning (DL) applications in earthquake engineering. The survey starts with an introduction and discussion of the methodological advances for elucidating several applicable DL techniques. A thorough research landscape is disclosed, examining the extent of DL applications across various research topics, including vision-based seismic damage assessment and structural characterization, seismic demand and damage state prediction, seismic response history prediction, regional seismic risk assessment and community resilience, ground motion (GM) for engineering use, seismic response control, and the inverse problem of system/damage identification. The review delineates suitable DL techniques for different research topics, highlighting the superior performance of CNN for vision-based tasks, RNN for sequential data, RL for community resilience, and unsupervised learning for dealing with GMs. The paper also discusses opportunities and challenges for leveraging DL to further advance earthquake engineering research and practice. The scarcity of labeled datasets and the imbalance of seismic damage instances pose hurdles for DL training, emphasizing the urgent need for open-access multimodel big data, against which cutting-edge DL techniques need to be explored and applied. Additionally, concerns arise regarding the interpretability and trustworthiness of DL models in safety-critical applications. Efforts to enhance model interpretability and incorporate physics into DL are crucial (yet still challenging for real-world complex structural systems) to foster trust and understanding within the community. While DL also shows the potential to address uncertainty quantification in earthquake engineering, additional research is advocated to further advance this research frontier.

## Acknowledgments

The author would like to thank Xiaoyu Liu and Chunxiao Ning from McGill University for preparing an initial list of relevant publications for this review article.

post-disaster reconnaissance. *Structural Control and Health Monitoring* 2020; **27**(4): 1–15. DOI: 10.1002/stc.2507.

59. Miao Z, Ji X, Wu M, Gao X. Deep learning-based evaluation for mechanical property degradation of seismically damaged RC columns. *Earthquake Engineering and Structural Dynamics* 2023; **52**(8): 2498–2519. DOI: 10.1002/eqe.3749.

60. Chou JS, Karundeng MA, Truong DN, Cheng MY. Identifying deflections of reinforced concrete beams under seismic loads by bio-inspired optimization of deep residual learning. *Structural Control and Health Monitoring* 2022; **29**(4): 1–24. DOI: 10.1002/stc.2918.

61. Li Y, Cui P, Ye C, Marcato Junior J, Zhang Z, Guo J, *et al.* Accurate prediction of earthquake-induced landslides based on deep learning considering landslide source area. *Remote Sensing* 2021; **13**(17). DOI: 10.3390/rs13173436.

62. Hacıefendioğlu K, Başağa HB, Demir G. Automatic detection of earthquake-induced ground failure effects through Faster R-CNN deep learning-based object detection using satellite images. *Natural Hazards* 2021; **105**(1): 383–403. DOI: 10.1007/s11069-020-04315-y.

63. Nex F, Duarte D, Tonolo FG, Kerle N. Structural building damage detection with deep learning: Assessment of a state-of-the-art CNN in operational conditions. *Remote Sensing* 2019; **11**(23). DOI: 10.3390/rs11232765.

64. Kawabe K, Horie K, Inoguchi M, Matsuoka M, Torisawa K, Liu W, *et al.* Extraction of Story-Collapsed Buildings By the 2016 Kumamoto Earthquake Using Deep Learning. *17th World Conference on Earthquake Engineering* 2020: 1–9.

65. Wang Y, Jing X, Xu Y, Cui L, Zhang Q, Li H. Geometry-guided semantic segmentation for post-earthquake buildings using optical remote sensing images. *Earthquake Engineering and Structural Dynamics* 2023; **52**(11): 3392–3413. DOI: 10.1002/eqe.3966.

66. Ma H, Liu Y, Ren Y, Yu J. Detection of collapsed buildings in post-earthquake remote sensing images based on the improved YOLOv3. *Remote Sensing* 2020; **12**(1). DOI: 10.3390/RS12010044.

67. Bai Y, Sezen H, Yilmaz A. End-to-end deep learning methods for automated damage detection in extreme events at various scales. *Proceedings - International Conference on Pattern Recognition* 2020: 5736–5743. DOI: 10.1109/ICPR48806.2021.9413041.

68. Wang J, Lei Y, Yang X, Zhang F. A refinement network embedded with attention mechanism for computer vision based post-earthquake inspections of railway viaduct. *Engineering Structures* 2023; **279**: 115572. DOI: 10.1016/j.engstruct.2022.115572.

69. Ogunjinmi PD, Park SS, Kim B, Lee DE. Rapid Post-Earthquake Structural Damage Assessment Using Convolutional Neural Networks and Transfer Learning. *Sensors* 2022; **22**(9): 1–20. DOI: 10.3390/s22093471.

70. Adriano B, Yokoya N, Xia J, Miura H, Liu W, Matsuoka M, *et al.* Learning from multimodal and multitemporal earth observation data for building damage mapping. *ISPRS Journal of Photogrammetry and Remote Sensing* 2021; **175**(July 2020): 132–143. DOI: 10.1016/j.isprsjprs.2021.02.016.

71. Bai Y, Zha B, Sezen H, Yilmaz A. Engineering deep learning methods on automatic detection of damage in infrastructure due to extreme events. *Structural Health Monitoring* 2023; **22**(1): 338–352. DOI: 10.1177/14759217221083649.

72. Bernabe S, Gonzalez C, Fernandez A, Bhangale U. Portability and Acceleration of Deep Learning Inferences to Detect Rapid Earthquake Damage from VHR Remote Sensing Images Using Intel OpenVINO Toolkit. *IEEE Journal of Selected Topics in Applied Earth Observations and Remote Sensing* 2021; **14**: 6906–6915. DOI: 10.1109/JSTARS.2021.3075961.

73. Bhatta S, Dang J. Multiclass seismic damage detection of buildings using quantum convolutional neural network. *Computer-Aided Civil and Infrastructure Engineering* 2023: 1–18. DOI: 10.1111/mice.13084.

74. Chen F, Yu B. Earthquake-Induced Building Damage Mapping Based on Multi-Task Deep Learning Framework. *IEEE Access* 2019; **7**: 181396–181404. DOI: 10.1109/ACCESS.2019.2958983.

75. Ci T, Liu Z, Wang Y. Assessment of the degree of building damage caused by disaster using convolutional neural networks in combination with ordinal regression. *Remote Sensing* 2019; **11**(23). DOI: 10.3390/rs11232858.

76. Fujita S, Hatayama M. Estimation Method for Roof-damaged Buildings from Aero-Photo Images During Earthquakes Using Deep Learning. *Information Systems Frontiers* 2021: 351–363. DOI: 10.1007/s10796-021-10124-w.

77. Gao Y, Mosalam KM. Deep Transfer Learning for Image-Based Structural Damage Recognition. *Computer-Aided Civil and Infrastructure Engineering* 2018; **33**(9): 748–768. DOI: 10.1111/mice.12363.

78. Hong Z, Yang Y, Liu J, Jiang S, Pan H, Zhou R, *et al.* Enhancing 3D Reconstruction Model by Deep Learning and Its Application in Building Damage Assessment after Earthquake. *Applied Sciences (Switzerland)* 2022; **12**(19). DOI: 10.3390/app12199790.

79. Ji M, Liu L, Du R, Buchroithner MF. A comparative study of texture and convolutional neural network features for detecting collapsed buildings after earthquakes using pre- and post-event satellite imagery. *Remote Sensing* 2019; **11**(10). DOI: 10.3390/rs11101202.

80. Ji M, Liu L, Zhang R, Buchroithner MF. Discrimination of earthquake-induced building destruction from space using a pretrained CNN model. *Applied Sciences (Switzerland)* 2020; **10**(2). DOI: 10.3390/app10020602.

81. Ji X, Zhuang Y, Miao Z, Cheng Y. Vision-based seismic damage detection and residual capacity assessment for an RC
22